\documentclass[conference]{IEEEtran}
\IEEEoverridecommandlockouts
\usepackage{cite}
\usepackage{amsmath,amssymb,amsfonts}
\usepackage{algorithmic}
\usepackage{graphicx}
\usepackage{textcomp}
\usepackage{xcolor}

\usepackage{makecell}
\usepackage{graphicx}
\usepackage{siunitx}
\usepackage{url}
\usepackage{hhline}
\usepackage{orcidlink}
\usepackage{booktabs}
\usepackage{algorithm}
\usepackage{algorithmic}

\def\BibTeX{{\rm B\kern-.05em{\sc i\kern-.025em b}\kern-.08em
    T\kern-.1667em\lower.7ex\hbox{E}\kern-.125emX}}
\begin{document}

\title{
Geometry-Preserving Aggregation for Mixture-of-Experts Embedding Models
}

\author{
\IEEEauthorblockN{Sajjad Kachuee \orcidlink{0000-0002-8187-6906}}
\IEEEauthorblockA{\textit{Department of Electrical Engineering} \\
\textit{Sharif University of Technology}\\
Tehran, Iran \\
sajjad.kachuee@gmail.com}
\and
\IEEEauthorblockN{Mohammad Sharifkhani \orcidlink{0009-0004-8609-7815}}
\IEEEauthorblockA{\textit{Department of Electrical Engineering} \\
\textit{Sharif University of Technology}\\
Tehran, Iran \\
msharifk@sharif.edu}
}


\maketitle

\begin{abstract}
Mixture-of-Experts (MoE) embedding models combine expert outputs using weighted linear summation, implicitly assuming a linear subspace structure in the embedding space. This assumption is shown to be inconsistent with the geometry of expert representations. Geometric analysis of a modern MoE embedding model reveals that expert outputs lie on a shared hyperspherical manifold characterized by tightly concentrated norms and substantial angular separation. Under this geometry, linear aggregation induces inward collapse toward the manifold interior, distorting vector magnitude and direction and reducing embedding comparability. To address this inconsistency, \emph{Spherical Barycentric Aggregation (SBA)} is introduced as a geometry-preserving aggregation operator that separates radial and angular components to maintain hyperspherical structure while remaining fully compatible with existing routing mechanisms. Experiments on selected tasks from the Massive Text Embedding Benchmark (MTEB), including semantic similarity, clustering, and duplicate question detection, demonstrate consistent performance improvements with identical training cost and full stability. Additional geometric analyses confirm that SBA prevents aggregation-induced collapse and preserves hyperspherical consistency, highlighting the importance of geometry-aware aggregation in MoE embedding architectures.
\end{abstract}


\section{Introduction}

Learning semantically meaningful vector representations is central to modern natural language processing (NLP). Embedding models underpin tasks such as semantic search, clustering, retrieval, and similarity estimation, where geometric comparability between representations is essential. In these settings, both vector magnitude and direction encode semantic information, and preservation of embedding-space structure directly impacts downstream performance. Methods such as Sentence-BERT \cite{reimers2019sentence} and large-scale embedding benchmarks such as MTEB \cite{muennighoff2023mteb} highlight the importance of stable and comparable embedding geometries.

In parallel, Mixture-of-Experts (MoE) architectures have emerged as an effective mechanism for scaling neural networks through sparse activation \cite{shazeer2017outrageously, fedus2022switch, lepikhin2020gshard}. By routing each input to a small subset of expert modules, MoE models increase parameter capacity without proportional increases in computational cost. Recent work has extended MoE architectures to embedding models \cite{nussbaum2024nomic, nussbaum2025trainingsparsemixtureexperts}, demonstrating strong retrieval and multilingual performance while maintaining efficiency.

Despite these advances, limited attention has been given to the geometric implications of expert aggregation in MoE embedding models. Standard MoE layers combine expert outputs using a weighted linear summation. While effective for predictive tasks, this operation implicitly assumes that expert representations reside in compatible linear subspaces and that linear interpolation preserves semantic structure. For embedding models—where cosine similarity and angular comparisons are fundamental—this assumption is nontrivial and warrants closer examination.

Empirical analysis reveals that MoE embedding experts do not behave as arbitrary linear components. Instead, expert outputs lie on a shared hyperspherical manifold characterized by tightly concentrated norms and substantial angular separation between experts. Under this geometry, specialization is encoded primarily in directional differences rather than magnitude.

Under a hyperspherical representation structure, linear weighted aggregation induces systematic inward collapse toward the interior of the manifold. This collapse alters both magnitude and direction, thereby violating geometric consistency of the embedding space. Consequently, the standard linear aggregation mechanism is structurally misaligned with the intrinsic geometry of expert representations.

To address this mismatch, \emph{Spherical Barycentric Aggregation (SBA)} is introduced as a geometry-preserving alternative to linear aggregation. SBA decomposes expert outputs into radial and angular components and performs aggregation directly on the hypersphere, ensuring that combined representations remain consistent with the underlying manifold. The proposed operator is lightweight, requires no modification to routing mechanisms or expert architectures, and integrates seamlessly into existing MoE frameworks.

Extensive experiments on standard embedding benchmarks demonstrate consistent improvements in similarity and clustering performance without additional computational cost or training instability. Complementary geometric analyses further validate that SBA prevents inward collapse and preserves hyperspherical consistency.

The main contributions are summarized as follows:

\begin{itemize}
    \item Identification of a geometric inconsistency between linear aggregation and hyperspherical expert representations in MoE embedding models.
    \item Empirical evidence demonstrating that MoE embedding experts operate on a shared hyperspherical manifold.
    \item Introduction of Spherical Barycentric Aggregation (SBA), a geometry-preserving alternative to linear aggregation.
    \item Consistent improvements on semantic similarity and clustering benchmarks with identical training cost and full training stability.
    \item Geometric validation and ablation studies clarifying the importance of radial--angular decomposition.
\end{itemize}

\section{Related Work}

\subsection{Scalable Mixture-of-Experts Architectures}

Mixture-of-Experts (MoE) architectures were introduced to scale neural networks through conditional computation \cite{shazeer2017outrageously}. By activating only a subset of experts per input, MoE layers increase parameter capacity without proportional computational cost. Subsequent work integrated sparse expert routing into large-scale Transformer architectures \cite{lepikhin2020gshard, fedus2022switch}, demonstrating improved efficiency and scalability. Extensions to vision models \cite{riquelme2021scaling} and dense-to-sparse training strategies \cite{pan2024dense} further established MoE as a dominant paradigm for scaling deep networks.

\subsection{Routing and Sparse Activation}

Routing mechanisms are central to MoE performance. Most modern systems employ softmax-based top-$K$ gating, where a learned router assigns tokens to a limited number of experts. Learning-to-route approaches \cite{dikkala2023benefits} demonstrate that trainable routing significantly improves performance over static assignments. Adaptive routing strategies \cite{zeng2024adamoe, yue2024ada} dynamically adjust expert utilization to balance efficiency and specialization. Additional work formulates routing as an optimization problem \cite{dong2025maximum} or introduces load-balancing objectives to prevent expert collapse \cite{guo2025advancing}.  

While these studies focus on expert selection efficiency and specialization, they do not examine the geometric properties of aggregated expert outputs.

\subsection{Aggregation of Expert Representations}

In contrast to routing, aggregation mechanisms in MoE architectures have received comparatively limited attention. The standard practice is weighted linear summation of selected expert outputs. This operation implicitly assumes that expert representations lie in compatible linear subspaces and that linear interpolation preserves semantic structure.

Recent work has explored alternatives to simple summation. For example, DAG-MoE \cite{anonymous2025dagmoe} introduces structured aggregation to model non-linear interactions between experts. However, existing approaches primarily aim to improve predictive performance and do not explicitly analyze the geometric consistency of embedding representations under aggregation.

\subsection{MoE for Embedding Models}

MoE architectures have recently been applied to embedding and retrieval models. Nomic Embed \cite{nussbaum2024nomic} demonstrates that sparse expert activation can produce competitive multilingual embeddings. Subsequent work on sparse embedding training \cite{nussbaum2025trainingsparsemixtureexperts} further explores efficiency–performance trade-offs in MoE-based representation learning. Other approaches, such as MoTE \cite{calvo2025mote}, allocate task-specific experts to improve cross-domain embedding comparability.

Although these works recognize that embedding outputs must remain comparable across inputs, they do not investigate whether standard aggregation preserves the intrinsic geometry of expert representations.

\subsection{Geometric Gap}

Across prior research, extensive effort has been devoted to scaling, routing efficiency, and expert specialization. However, the geometric structure of expert outputs in MoE embedding models—and the compatibility of linear aggregation with that structure—remains largely unexplored.  

In embedding models, where cosine similarity and angular relationships are fundamental, aggregation operators directly influence representation geometry. The absence of geometric analysis in prior MoE embedding work motivates the present study and the development of a geometry-preserving aggregation mechanism.

\section{Hyperspherical Geometry of MoE Expert Outputs}
\label{sec:geometry}

To investigate the intrinsic geometry of expert representations in MoE embedding models, the publicly available \texttt{nomic-embed-text-v2-moe} model \cite{nussbaum2025trainingsparsemixtureexperts} is analyzed. This state-of-the-art sparse MoE embedding architecture employs top-2 routing, activating two experts per token while maintaining a shared embedding dimensionality across experts.

Expert outputs are extracted from all MoE layers prior to aggregation, enabling direct geometric inspection of individual expert representations. Geometric statistics are computed over a large corpus of prompts sampled from the MMLU benchmark \cite{hendrycks2020measuring}, spanning 57 subject domains to ensure broad semantic coverage. All reported measurements are aggregated across layers, tokens, and tasks to characterize the latent representation manifold underlying MoE embedding models.

\subsection{Expert Output Norm Distribution}

\begin{figure}[ht]
\centering
\includegraphics[width=1.0\linewidth]{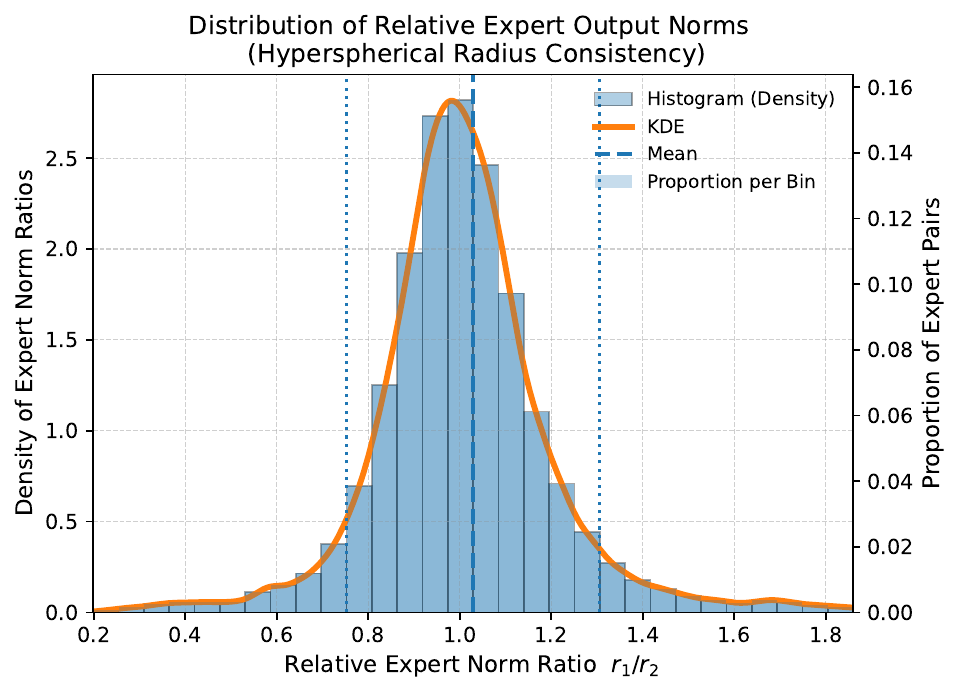}
\caption{
Distribution of relative expert output norm ratios ($r_1 / r_2$) across all MoE layers and MMLU skill categories. The tight concentration around 1 demonstrates that active experts produce embeddings with nearly identical radial magnitude, indicating that expert representations lie on a shared hyperspherical manifold.
}
\label{fig:norm_distribution}
\end{figure}

For each routed expert output $e_i(x)$, the Euclidean norm $\|e_i(x)\|$ is computed across all MoE layers and input samples. Figure~\ref{fig:norm_distribution} shows the distribution of relative norm ratios between active expert pairs (top-2 routing).

The relative norm ratios are tightly concentrated around unity, indicating that expert outputs exhibit highly consistent magnitudes across layers and tasks. This radial consistency suggests that expert representations do not occupy independent linear subspaces with varying scales. Instead, the outputs concentrate around a common radius, consistent with a shared hyperspherical embedding structure.

\subsection{Angular Separation Between Experts}

In addition to radial consistency, the angular relationships between simultaneously activated experts are examined. For each active expert pair $(e_i, e_j)$, the pairwise angular distance is computed.

\begin{figure}[ht]
\centering
\includegraphics[width=1.0\linewidth]{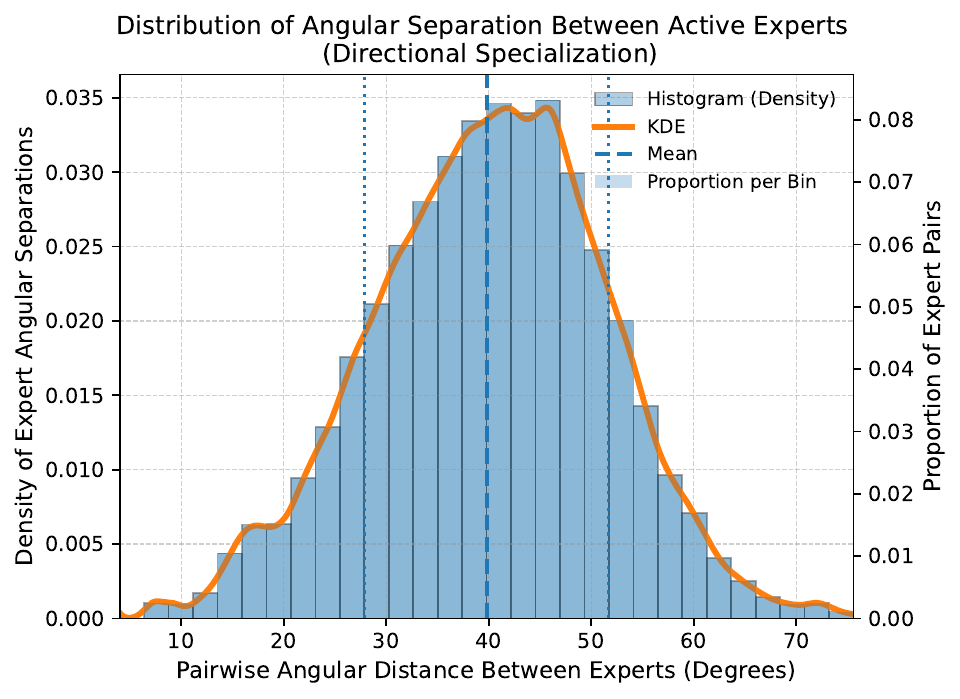}
\caption{
Distribution of pairwise angular distances between active expert outputs across all MoE layers and MMLU skill categories. The majority of angular separations exceed $40^\circ$, demonstrating strong directional specialization among experts despite shared radial magnitude.
}
\label{fig:angle_distribution}
\end{figure}

As shown in Figure~\ref{fig:angle_distribution}, most pairwise angles exceed $40^\circ$, indicating substantial directional separation between experts. This demonstrates that specialization among experts is primarily encoded through directional variation rather than differences in magnitude.

The combination of near-constant norms and large angular separation strongly indicates that MoE embedding experts operate on a shared hyperspherical manifold, where semantic differentiation is expressed through angular structure.

\section{Geometry-Preserving Aggregation}

\subsection{Hyperspherical Structure of Expert Representations}

Section~III established that MoE embedding experts exhibit two consistent geometric properties: (i) tightly concentrated norms and (ii) substantial angular separation between simultaneously activated experts. These observations indicate that expert outputs lie on a shared hyperspherical manifold, where semantic specialization is primarily encoded through directional variation rather than magnitude differences.

Under such geometry, aggregation mechanisms should preserve both radial consistency and angular structure in order to maintain embedding comparability.

\subsection{Geometric Failure of Linear Aggregation}

In standard MoE layers, selected expert outputs are combined using weighted linear summation:

\begin{equation}
y_{\text{linear}} = \sum_{i \in \text{Top-}K} w_i e_i .
\end{equation}

Assuming comparable expert norms $\|e_i\| \approx r$, the magnitude of the aggregated output becomes

\begin{equation}
\|y_{\text{linear}}\| 
= r \left\| \sum_i w_i \hat{e}_i \right\| ,
\end{equation}

where $\hat{e}_i = e_i / \|e_i\|$ denotes the unit-normalized direction.

Unless all expert directions are perfectly aligned, the angular diversity among experts produces partial cancellation in the weighted summation, yielding

\begin{equation}
\|y_{\text{linear}}\| < r .
\end{equation}

As a result, linear aggregation systematically shifts representations toward the interior of the hypersphere. This inward collapse alters both magnitude and direction, violating the intrinsic geometric structure of the embedding space.

\begin{figure}[t]
\centering
\includegraphics[width=\linewidth]{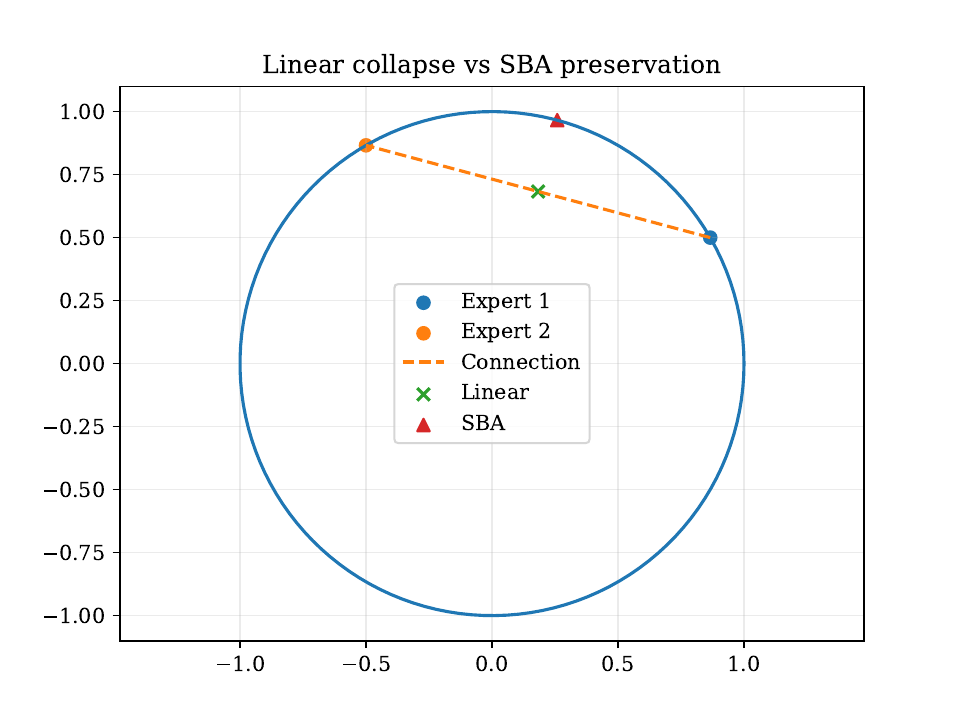}
\caption{Conceptual illustration of geometric collapse under linear aggregation. Two expert embeddings lie on a shared hypersphere with large angular separation. Linear weighted aggregation collapses inward, while Spherical Barycentric Aggregation (SBA) preserves hyperspherical geometry.}
\label{fig:conceptual}
\end{figure}

Figure~\ref{fig:conceptual} conceptually illustrates the geometric inconsistency of linear aggregation and the manifold-preserving behavior of SBA.

\subsection{Spherical Barycentric Aggregation (SBA)}

\begin{figure}[t]
\centering
\includegraphics[width=\linewidth]{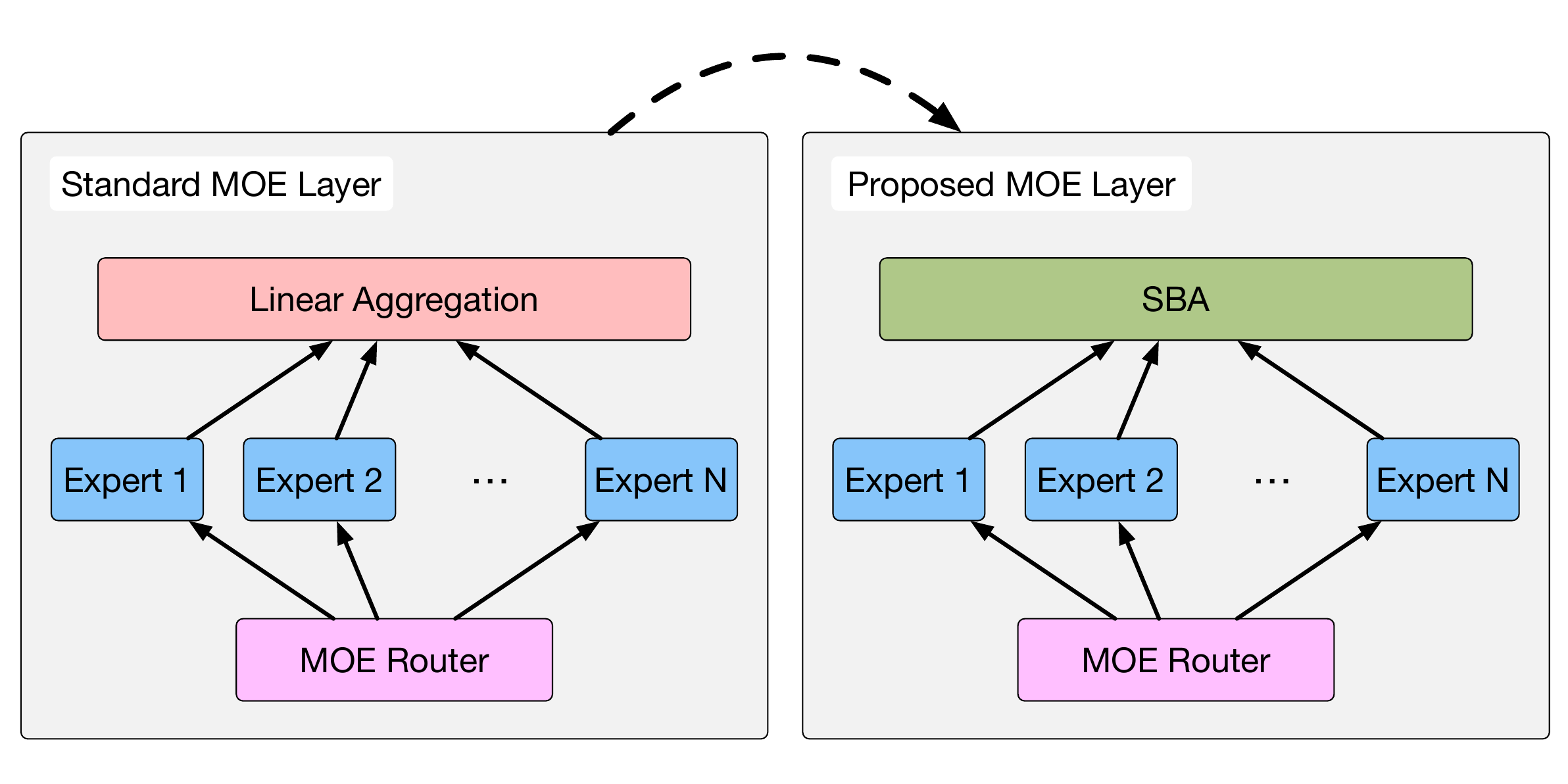}
\caption{Integration of Spherical Barycentric Aggregation (SBA) into a standard Mixture-of-Experts layer. SBA replaces linear weighted summation with a geometry-preserving aggregation operator while leaving routing, expert computation, and training procedures unchanged.}
\label{fig:sba_layer}
\end{figure}

To preserve hyperspherical consistency, \emph{Spherical Barycentric Aggregation (SBA)} is introduced as a geometry-aware alternative to linear summation. The method separates radial and angular components prior to aggregation. Figure~\ref{fig:sba_layer} illustrates how SBA replaces linear aggregation within a standard MoE layer without modifying routing or expert computation.

For clarity, the formulation is first described for the top-2 routing case. Let the routed experts produce outputs $e_1, e_2$ with router weights $w_1, w_2$.

\paragraph{Radial--Angular Decomposition}

Each expert output is decomposed into magnitude and direction:

\begin{equation}
r_i = \|e_i\|, 
\qquad
\hat{u}_i = \frac{e_i}{\|e_i\|}.
\end{equation}

\paragraph{Radius Aggregation}

The aggregated radius is computed as the weighted barycenter of expert magnitudes:

\begin{equation}
r = w_1 r_1 + w_2 r_2 .
\end{equation}

\paragraph{Angular Aggregation}

The aggregated direction is computed using a norm-aware angular barycenter:

\begin{equation}
\theta =
\frac{w_1 r_1 \theta_1 + w_2 r_2 \theta_2}
     {w_1 r_1 + w_2 r_2}.
\end{equation}

This formulation ensures that experts with both higher router confidence and larger magnitude exert proportionally stronger influence on the resulting direction.

\paragraph{Reconstruction}

The final aggregated representation is reconstructed on the hypersphere:

\begin{equation}
y_{\text{SBA}} = r \cdot \hat{u}(\theta),
\end{equation}

where $\hat{u}(\theta)$ denotes the unit vector corresponding to angular coordinate $\theta$.

Unlike linear summation, SBA preserves hyperspherical structure while remaining fully compatible with existing routing mechanisms and training procedures.

\subsection{Generalization to Top-\texorpdfstring{$K$}{K} Experts}

For $K > 2$, the formulation generalizes naturally:

\begin{equation}
r = \sum_{i=1}^{K} w_i r_i,
\qquad
\theta =
\frac{\sum_{i=1}^{K} w_i r_i \theta_i}
     {\sum_{i=1}^{K} w_i r_i}.
\end{equation}

This generalization maintains hyperspherical consistency regardless of the number of activated experts.

\subsection{Ablation Variants}

To better understand the role of radial and angular components in SBA, two controlled ablation variants are evaluated.

\paragraph{Norm-Free Angular Aggregation}

This variant removes expert magnitude from the angular barycenter while preserving the radial aggregation step. The output radius is computed as in full SBA:

\begin{equation}
r = \sum_{i=1}^{K} w_i r_i ,
\end{equation}

but the angular component is computed using a pure weight-average:

\begin{equation}
\theta_{\text{norm-free}} =
\frac{\sum_{i=1}^{K} w_i \theta_i}
     {\sum_{i=1}^{K} w_i}.
\end{equation}

This ablation isolates the effect of incorporating expert norms into directional aggregation.

\paragraph{Unit-Normalized Output (Radius-Fixed)}

This variant preserves the norm-aware angular aggregation of SBA but removes the radial component during reconstruction. After computing the angular barycenter $\theta_{\text{SBA}}$, the output is explicitly normalized to unit length:

\begin{equation}
y_{\text{unit}} = \hat{u}(\theta_{\text{SBA}}).
\end{equation}

Equivalently, the aggregated radius is discarded, forcing all representations to lie on the unit hypersphere regardless of expert magnitudes.

This ablation evaluates whether preserving the aggregated radius contributes meaningful information to downstream tasks.

\section{Experimental Setup}

\subsection{Model and Architecture}

Experiments are conducted using the publicly available MoE embedding model \texttt{nomic-ai/nomic-embed-text-v2-moe} \cite{nussbaum2024nomic}. This model employs sparse top-$K$ routing with $K=2$ experts dynamically activated per token at each MoE layer.

Two variants are evaluated:

\begin{itemize}
    \item \textbf{Linear MoE (Baseline):} Standard weighted linear aggregation of expert outputs.
    \item \textbf{SBA MoE:} Linear aggregation replaced with Spherical Barycentric Aggregation (SBA), while keeping routing, expert parameters, and pretrained weights unchanged.
\end{itemize}

The modification is strictly limited to the aggregation operator.

\subsection{Training Configuration}

All models are fine-tuned under identical optimization settings to ensure fair comparison. Training is performed using PyTorch and HuggingFace Transformers on a single NVIDIA T4 GPU.

Fine-tuning is conducted for three epochs on 30K triplets using AdamW with a learning rate of $2 \times 10^{-6}$. A linear learning-rate scheduler with 10\% warmup is applied. Gradient accumulation is used to achieve an effective batch size of 64.

The training objective is Multiple Negatives Ranking Loss (InfoNCE) \cite{reimers2019sentence, henderson2017efficient}:

\begin{equation}
\mathcal{L} = - \frac{1}{N} 
\sum_{i=1}^{N}
\log
\frac{e^{\text{sim}(a_i, p_i)/\tau}}
     {\sum_{j=1}^{N} e^{\text{sim}(a_i, p_j)/\tau}},
\end{equation}

where $\text{sim}(\cdot,\cdot)$ denotes cosine similarity and $\tau = 0.05$.

\subsection{Datasets}

\paragraph{Training Data}

Fine-tuning uses the \texttt{sentence-transformers/all-nli} dataset \cite{reimers2019sentence}, which combines SNLI \cite{bowman2015large} and MultiNLI \cite{williams2018broad} sentence pairs for representation learning.

\paragraph{Evaluation Data}

Evaluation is performed on selected tasks from the Massive Text Embedding Benchmark (MTEB) \cite{muennighoff2023mteb}:

\begin{itemize}
    \item \textbf{STSBenchmark:} Semantic similarity evaluated using Spearman correlation \cite{cer2017semeval}.
    \item \textbf{StackExchangeClustering:} Clustering evaluated using V-measure \cite{geigle2021tweac}.
    \item \textbf{SprintDuplicateQuestions:} Pair classification evaluated using cosine-based Average Precision \cite{shah2018adversarial}.
\end{itemize}

\subsection{Evaluation Protocol}

Sentence embeddings are extracted from the final MoE layer; for SBA models, embeddings are produced via spherical barycentric aggregation of expert outputs. Pairwise embedding similarity is computed using cosine similarity, and task-specific evaluation metrics follow the standard MTEB protocol described in the previous subsection. All reported results are averaged over three random seeds to ensure stability.

\section{Results}

\subsection{Main Benchmark Results}

Table~\ref{tab:main_results} compares the baseline Linear MoE with the proposed SBA MoE.

SBA consistently improves performance across all evaluated tasks. The largest improvement is observed on SprintDuplicateQuestions, where performance increases by more than five points, suggesting that geometry-preserving aggregation enhances pairwise semantic discrimination. Additional gains in clustering and semantic similarity indicate consistent benefits across diverse embedding settings.

\begin{table}[t]
\centering
\caption{Comparison of Linear MoE and SBA MoE on selected MTEB tasks.}
\label{tab:main_results}
\begin{tabular}{lccc}
\hline
Method & STSBenchmark & StackExchange & SprintDuplicate \\
\hline
Linear MoE & 71.18 & 85.66 & 83.66 \\
SBA MoE & \textbf{71.51} & \textbf{86.19} & \textbf{89.06} \\
\hline
\end{tabular}
\end{table}

\subsection{Geometric Validation}

To quantify the geometric effect of aggregation, the relative norm ratio $\|y\|/\bar{r}$ is measured, where $\bar{r} = (r_1 + r_2)/2$ denotes the average norm of the contributing experts in a top-2 routed layer. This ratio directly measures deviation from the hyperspherical expert manifold.

\begin{figure}[t]
\centering
\includegraphics[width=\linewidth]{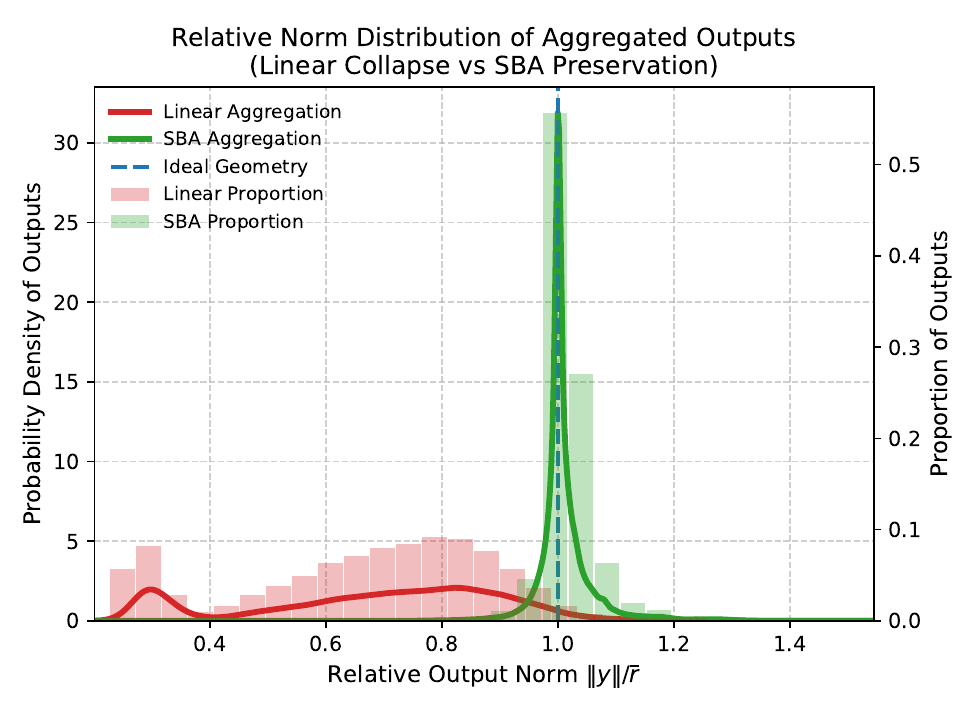}
\caption{Distribution of relative output norm ratios $\|y\|/\bar{r}$ for linearly aggregated and SBA-aggregated embeddings. Linear aggregation exhibits inward collapse toward the hypersphere center, whereas SBA preserves hyperspherical magnitude and maintains manifold consistency.}
\label{fig:collapse}
\end{figure}

As shown in Figure~\ref{fig:collapse}, linear aggregation produces outputs with systematically reduced norm, indicating inward collapse due to directional cancellation among experts. In contrast, SBA maintains relative norms tightly concentrated around unity, confirming preservation of hyperspherical structure and explaining the observed downstream performance gains.

\subsection{Ablation Study}

Table~\ref{tab:ablation} reports results for the two ablation variants.

\begin{table}[t]
\centering
\caption{Ablation results for SBA variants.}
\label{tab:ablation}
\begin{tabular}{lccc}
\hline
Method & STSBenchmark & StackExchange & SprintDuplicate \\
\hline
Linear & 71.18 & 85.66 & 83.66 \\
SBA (full) & \textbf{71.51} & \textbf{86.19} & 89.06 \\
Norm-free angle & 71.37 & 85.64 & \textbf{89.69} \\
Unit-normalized & 68.35 & 0.00 & 88.19 \\
\hline
\end{tabular}
\end{table}

Removing norm information from the angular barycenter slightly degrades performance on most tasks, indicating that expert magnitude contributes to directional aggregation. Interestingly, on SprintDuplicateQuestions the norm-free variant slightly outperforms full SBA. This task primarily relies on fine-grained pairwise ranking based on cosine similarity, and pure angular averaging may preserve sharper directional contrast in certain cases. However, the norm-free formulation underperforms on clustering and similarity benchmarks, suggesting that incorporating radial information improves general embedding robustness across diverse evaluation settings.

Forcing unit normalization significantly degrades clustering performance, indicating that the aggregated radius contains discriminative information that is important for structured embedding spaces. These results collectively suggest that coupling radial and angular components provides the most consistent overall performance.

\section{Discussion and Limitations}

The analysis and experiments indicate that MoE embedding experts operate on a shared hyperspherical manifold in which specialization is primarily encoded through directional variation rather than magnitude. Under this geometric structure, weighted linear aggregation introduces systematic inward collapse, altering both vector magnitude and direction. This distortion reduces geometric consistency and can negatively affect embedding comparability.

Spherical Barycentric Aggregation (SBA) addresses this mismatch by preserving both radial and angular structure during aggregation. By maintaining hyperspherical consistency, SBA produces embeddings that remain geometrically comparable to individual expert outputs. The observed improvements in similarity and clustering tasks suggest that preserving manifold structure leads to more regular and transferable embedding spaces.

Importantly, SBA is lightweight and requires no modification to routing mechanisms, expert architectures, or training procedures. This makes the method directly applicable to existing MoE embedding systems and potentially beneficial for other representation-learning settings where cosine similarity and angular relationships are fundamental.

\subsection*{Limitations}

Several limitations should be considered. First, experiments are conducted on a single MoE embedding architecture; evaluating additional models would further validate generality. Second, SBA assumes that expert outputs follow a hyperspherical structure. Although this assumption is empirically supported in our analysis, different normalization strategies or training objectives may alter representation geometry. Third, the present study focuses on top-$K=2$ routing; interactions between SBA and larger expert sets remain an open question.

\subsection*{Broader Implications}

The findings suggest that respecting embedding-space geometry is important when designing aggregation operators for sparse expert models. Geometry-aware aggregation may benefit other MoE-based systems, multimodal representation learning, and large-scale retrieval architectures where angular similarity governs downstream performance.

\section{Conclusion}

This work examined the geometric structure of expert representations in Mixture-of-Experts (MoE) embedding models and identified a fundamental mismatch between hyperspherical expert geometry and standard linear aggregation. Empirical analysis demonstrated that expert outputs exhibit tightly concentrated norms and substantial angular separation, forming a shared hyperspherical manifold. Under this structure, weighted linear summation induces systematic inward collapse, distorting embedding magnitude and direction.

To address this inconsistency, Spherical Barycentric Aggregation (SBA) was introduced as a geometry-preserving alternative that separates radial and angular components during aggregation. SBA maintains hyperspherical consistency while remaining fully compatible with existing routing mechanisms and training procedures. Experimental results on multiple embedding benchmarks confirmed consistent performance improvements without additional computational cost.

These findings underscore the importance of respecting embedding-space geometry when designing aggregation operators for sparse expert architectures. Geometry-aware aggregation offers a principled direction for improving MoE-based representation learning systems and may extend to broader retrieval and multimodal embedding frameworks.

\section*{Acknowledgment}

The authors thank Hamidreza Bandeali, Mohammad Kolahdouzan, and Amirreza Farahbakhsh for helpful discussions and insightful feedback during the development of this work.

\bibliographystyle{IEEEtran}
\bibliography{custom.bib}

\end{document}